\documentclass{ecai}
\usepackage{graphicx}
\usepackage{latexsym}
\usepackage{wasysym}
\usepackage{hyperref}

\begin{document}

\begin{frontmatter}

\title{Cognitive Effects in Large Language Models}

\author[A]{\fnms{Jonathan}~\snm{Shaki}\thanks{Corresponding Author. Email: jonath.shaki@gmail.com.}}

%\author[A]{\fnms{First}~\snm{Author}\orcid{....-....-....-....}\thanks{Corresponding Author. Email: somename@university.edu.}}
%\author[B]{\fnms{Second}~\snm{Author}\orcid{....-....-....-....}}
%\author[B]{\fnms{Third}~\snm{Author}\orcid{....-....-....-....}} % use of \orcid{} is optional

\author[A]{\fnms{Sarit}~\snm{Kraus}}
\author[B]{\fnms{Michael}~\snm{Wooldridge}}

\address[A]{Department of Computer Science, Bar-Ilan University, Israel}
\address[B]{Department of Computer Science, University of Oxford, United Kingdom}

\begin{abstract}
Large Language Models (LLMs) such as ChatGPT have received enormous attention over the past year and are now used by hundreds of millions of people every day. The rapid adoption of this technology naturally raises questions about the possible biases such models might exhibit. In this work, we tested one of these models (GPT-3) on a range of cognitive effects, which are systematic patterns that are usually found in human cognitive tasks. We found that LLMs are indeed prone to several human cognitive effects. Specifically, we show that the \emph{priming}, \emph{distance}, \emph{SNARC}, and \emph{size congruity} effects were presented with GPT-3, while the \emph{anchoring} effect is absent. We describe our methodology, and specifically the way we converted real-world experiments to text-based experiments. Finally, we speculate on the possible reasons why GPT-3 exhibits these effects and discuss whether they are imitated or reinvented. 
\end{abstract}
\end{frontmatter}

\section{Introduction}

Over the past year, Large Language Models have established themselves as the most influential and widely-adopted AI technology to date. Such models are very large-scale machine learning models designed to process and produce realistic human text. The standard way in which these models are trained is to present them with essentially all the digital text available in the world -- this text is typically scraped from the World-Wide Web, social media, and indeed every other source of digital text that the developers can obtain. Because of their inherently opaque nature, concerns have been raised about possible biases and hidden goal structures that such models may acquire. For example, there is a huge amount of racist and misogynistic content on the World-Wide Web, and while designers might try to filter out the most obviously toxic content, the scale of the training data means this must inevitably be an imperfect process. Moreover, humans fall prey to endless unconscious biases, and it seems inevitable, firstly, that these will be reflected in the training data, and, second, that the models will then acquire these biases themselves. It is important, therefore, to understand the extent to which this occurs, so that we can identify it, and hopefully remedy it. A deeper understanding of these issues will, in turn lead to a deeper understanding of the nature and operation of the technology itself -- which at present remains rather poorly understood.

This work aims to further our understanding of these issues. We investigate the extent to which Large Language Models (GPT-3 in particular) exhibit the \emph{cognitive effects} that human cognitive processes exhibit. Such effects have been widely studied within cognitive psychology. We begin by describing LLMs and cognitive effects in more detail; we then describe our methodology and then investigate the extent to which GPT-3 exhibits five key cognitive effects. We conclude with a discussion.

\noindent\textbf{Large Language Models:}
In more detail, \emph{Large Language Models} (LLMs) are deep learning models that have been trained to produce human-like text, typically by computing a probability distribution over the possible completions of a given sequence of tokens. Typically, an LLM is fed by token sequences, where tokens represent multiple characters in an embedding space~\cite{radford2018improving}. With this technique, words with similar meaning are closer in the embedding space, hence have more similar representation than unrelated words. This is intended to enhance the performance of pre-trained models by making it more natural to the model to treat similar words in similar ways \cite{almeida2019word}.
GPT-3 is a transformer based pre-trained language model \cite{floridi2020gpt}. Before being fed into GPT-3, the text is converted into pre-embedded words (tokens), which are then fed to GPT-3. As with most language models, for each token sequence (\emph{prompt}) GPT-3 computes a probability distribution for the next token. In order to produce longer texts, tokens are chosen from this distribution by some criterion and then fed back into GPT-3 as input tokens.

Transformers are deep learning models that are usually used for sequence-to-sequence prediction, such as text generation. They have demonstrated remarkable performance on some important tasks, and have some interesting characteristics, among them the self-attention mechanism, which allows such models to concentrate on specific parts of the input sequence \cite{lin2022survey}. This is an important consideration for our work, as GPT-3's capacity to selectively attend to certain segments of its input sequence, due to its transformer-based architecture, could account for certain outcomes. % In addition, if their activity was accessible through Open-AI API, it could also provide a way to explore more aspects of the tested cognitive effects.

\vspace*{1.5ex}\noindent\textbf{Cognitive effects:}
Cognitive processes are mental constructs such as memory, perception, learning, reasoning, and recall. Investigation of these processes is the main method through which experimental psychologists try to understand the way the brain processes information. Through examination of cognitive effects, which are the expressions of systematic patterns in human cognition, psychologists can learn about these processes in various domains \cite{barsalou2014cognitive,anderson1996architecture}.

For example, the \emph{priming effect} refers to the influence of a stimulus presented on the subsequent processing of another stimulus. In a typical experimental setting, participants are required to decide whether a target letter sequence forms a word. Participants respond faster when target words follow semantically-associated words than unrelated words \cite{sperber1979semantic,janiszewski2014content}. The standard explanation is that the context made by the prime stimulus makes the cognitive operations used to comprehend this content more accessible \cite{janiszewski2014content}. 

An example from the literature of comparative judgement is the \emph{distance effect}. In various experiments, participants were required to compare the size of two presented stimuli, typically numbers or animals. The reaction times decreased as the difference between the stimuli' sizes increased \cite{moyer1967time,moyer1973comparing,dehaene1998abstract}. This is usually attributed to a ``mental number line'' on which comparisons are done, such that it is easier to compare/encode distant points than close points \cite{dehaene1992varieties}.

A small number of studies that test cognitive biases in GPT-3 have previously appeared. \cite{binz2023using} tested cognitive biases regarding rationality and judgement, and found some human-like errors that GPT-3 tends to make. Another study \cite{jones2022capturing} tested similar biases, including the anchoring effect, which we were unable to replicate (see below). \cite{azariachatgpt} found that GPT-3 tends to generate digits with a distribution similar to that of humans. Finally, \cite{park2023artificial} tried to replicate 14 experiments that were done on human participants with GPT-3, mostly including moral and rational dilemmas. They successfully replicated only a few of them, however, mainly due to what they call ``the correct answer effect''. According to the study, in six of the 
experiments, GPT-3 responded with near zero variance between different conditions due to high certainty in a specific answer (though sometimes subjectively true, e.g., political opinions). In some of our experiments we encountered a similar problem -- see the general methodology section, below, for details.

In contrast to these previous studies, we focus less on rationality and judgement and mostly investigate effects relating to mental processes. In doing so, we hope to shed more light on the way GPT-3's ``cognition'' works, in the same way that the study of cognitive effects shed light on human cognition. An important question we raise here is whether these effects are \emph{imitated} or \emph{reinvented} by GPT-3. By examining diverse cognitive effects from different literature, we hope to provide insights to this question. Finally, note that if these effects are indeed reinvented, they might contribute to the field of psychology itself, by providing examples of these effects beyond the realm of the human cognition.

\section{A Note on Methodology}

LLMs are a fundamentally new type of AI system – large, complex, and inherently opaque. Interaction with them take the form of a natural language dialogue, rather than via a language with well-defined formal semantics (such as logic). In addition, for LLMs to be useful, it is often necessary to introduce an element of randomness into their responses (e.g., via the ``temperature'' parameter in GPT-3). For these reasons, many attempts to systematically evaluate LLMs have failed because of flaws in the evaluation methodology. For this reason, and in the interests of reproducibility, we here comment on our approach. 

The majority of the effects in cognitive psychology were originally discovered by comparing the reaction time of different conditions, apparently reflecting the difficulty of the task (so-called \emph{mental chronometry}~\cite{posner2005timing}). In contrast, the response times of an LLM are a function of the length of the prompt, regardless of its content. Hence, we measure GPT-3's confidence, the probability it assigns to the correct prediction, in proportion to the overall probability assigned to relevant predictions. This is analogous to error rate, the accompanied dependent variable of reaction time, used to identify the task's difficulty.

When using confidence as the dependent variable, it is important to note that confidence can approach 1, and then it is not possible to measure possible differences between conditions (this was identified but not resolved in \cite{park2023artificial}). In order to overcome this methodological problem, we added ``mental load'' to make the task harder, typically by adding spaces between the stimulus's letters. We have ignored queries for which non of the 5 top probabilities were assigned to a relevant answer, which rarely happened.

There are some limitations regarding the manipulation of the independent variables. For example, cognitive biases in human participants were found by comparing responses to clear versus blurred text \cite{roediger1987effects} or small versus large fonts \cite{rubinsten2002ant}. Other cognitive effects were found by comparing responses with left versus right hands \cite{dehaene1993mental}. None of these manipulations can be done directly with GPT-3. In order to overcome these technical limitations, we implemented the following original solutions. For example, we ``blurred'' text by adding spaces between letters and presented words with different capitalization as font size analogy. In addition, we have directed GPT-3 to respond with the words ``left'' and ``right'', instead of spatial responses.    
Finally, human cognitive biases were found by testing many participants and averaging their data. In contrast, exploring GPT-3's biases means testing one participant only, as GPT-3 always responds with an identical probability distribution for identical queries (``deterministic probability''). In order to overcome this statistical-power problem ($N=1$), we have asked the same question with multiple formats to address the issue of having only one participant. In this way, we were able to ask multiple queries and obtain sufficient data for analysis.

Although these modifications make our experiments slightly different from traditional methodologies, we claim that our novel methodology capture the essence of the various effects. The fact that we have been able to replicate most of the cognitive effects we tested supports this claim.

All the experiments were done with GPT-3, and specifically on the model \texttt{text-davinci-003}.

\section{Our Study}

We investigated five cognitive effects from various fields. We present the effects, the methodologies used in our experiments, and the findings. The priming, distance, SNARC, and size congruity effects were detected in GPT, while the anchoring effect was absent.

\subsection{The Priming Effect}

The priming effect pertains to the influence of a stimulus presented (the \emph{priming stimulus}) on the subsequent processing of another stimulus (the \emph{target stimulus}). Such influence is typically attributed to make similar concepts more available \cite{janiszewski2014content}. Similar to experiments done with human participants, our objective was to determine whether GPT-3's ability to recognize words is improved when they are presented after associated words, in comparison to when they are presented after unrelated words \cite{sperber1979semantic,janiszewski2014content,hutchison2013semantic}.

We asked GPT-3 to complete a prompt of one of the three following formats:
\begin{enumerate}
    \item 
The ``question'' variation: \\
Q: Answer with an arbitrary word. \\
A: {[prime stimulus]}. \\
Q: Can the letter sequence "[target stimulus]" form a word? \\
A: 
\item The ``sentence'' variation: \\
"[prime stimulus]" is a word. \\
Q: Can the letter sequence "[target stimulus]" form a word? \\
A: 
\item The ``simple'' variation: \\
{[prime stimulus]}. \\
Q: Can the letter sequence "[target stimulus]" form a word? \\
A: 
\end{enumerate}
The target stimulus was presented in two different conditions, using an unrelated word or an associated word as the prime stimulus. To ensure that recognizing the words was not too easy, we separated the letters of the target stimulus from each other, by 5, 10, or 15 spaces in different experiments.

Our stimuli were drawn from the semantic priming project \cite{hutchison2013semantic}, which contains a list of over 1000 target words, each with one associated word and two unrelated words. We selected 300 target words, including three sets of 100 words, of 4, 5, and 6 letter lengths. The criteria for selection were the words with the strongest association (as was tested empirically on humans) while having at least one unrelated word with an association weaker than 0.2, which served as the unrelated priming stimulus.

In order to test GPT-3 multiple times on the same stimulus, we introduced some variations on the query format. The question and answer indicators (in the example above, Q\&A) could be Q\&A or Question\&Answer. The separator between the indicator and the rest of text could be ``:'', ``)'', or ``.''. Every step was performed with all combinations (i.e., $2 \cdot 3 = 6$ combinations).

In addition, in order to ensure that GPT-3 understands the questions, we added 100 queries with catch trials of non-word targets. These targets were composed from 5-letter sequences without vowels, and presented with 15 spaces and the "question" variation.

\vspace*{1.5ex}\noindent\textbf{Results:}
As noted before, the confidence is the probability GPT-3 assigns to the correct answer (``yes'') in proportion to the combined probability of both ``yes'' and ``no'' answers. We omitted words that GPT-3 identifies with high certainty as words (above $.99$ confidence) for both relevant and non-relevant priming, as well as words that GPT-3 could not recognize (below $.6$ confidence) for both types of priming. Finally, we averaged the confidence over the number of spaces, for each word.
The results are summarised in the following table.
\begin{table}
\begin{center}
{\caption{The priming effect}\label{table1}}
\begin{tabular}{lccccccc}
\hline
\rule{0pt}{12pt}
Experiment&$\Diamond$&$\Box$&$\bigtriangleup$&$\bigcirc$&$\pentagon$&$\times$&
\\
\hline
\\[-6pt]
\quad 4-sentence & .56 & .76 & <0.001 & -9.78 & 946 & 79 \\
\quad 5-sentence & .62 & .81 & <0.001 & -10.19 & 1066 & 89 \\
\quad 6-sentence & .66 & .81 & <0.001 & -8.1 & 1030 & 86 \\
\quad 4-question & .78 & .85 & <0.001 & -4.83 & 1054 & 88 \\
\quad 5-question & .81 & .85 & 0.0044 & -2.85 & 970 & 81 \\
\quad 6-question & .81 & .85 & 0.004 & -2.89 & 958 & 80 \\
\quad 4-simple & .79 & .74 & 0.0054 & 2.79 & 1138 & 95 \\
\quad 5-simple & .79 & .78 & 0.726 & 0.35 & 1114 & 93 \\
\quad 6-simple & .78 & .76 & 0.3037 & 1.03 & 1114 & 93 \\
\\
\hline
\\[-6pt]
\multicolumn{8}{l}{
$\Diamond$ confidence with irrelevant priming \ \
$\Box$ with relevant priming \ \
}
\\
\multicolumn{8}{l}{
$\bigtriangleup$ p-value \ \
$\bigcirc$ t-statistic \ \
$\pentagon$ degrees of freedom
}
\\
\multicolumn{8}{l}{
$\times$ analyzed words \ \
}
\end{tabular}
\end{center}
\end{table}

The priming effect was present in the ``sentence'' and ``question'' variations, although stronger for the former. However, the effect was absent in the ``simple'' variation. This could imply that the transformers in GPT-3's attention mechanism may focus more on the primer when it is explicitly presented as a word, as it appears more relevant to the query. This mechanism can explain the impact of the priming words in the ``sentence'' and ``question'' variations. Alternatively, it is possible that the attention mechanism ignores words that are out of context, as in the ``simple'' variation.

For the non-words case the average confidence was higher than $.99$, that is, they were classified correctly by GPT-3 as non-words, and so we conclude GPT-3 indeed ``understood'' the question.

\subsection{The Distance Effect}

Another widely-accepted phenomenon in cognitive psychology is the distance effect, which suggests that the amount of time needed to compare two symbols is dependent on the distance between their referents along the dimension being evaluated. The effect has been consistently demonstrated across various domains, such as perception and symbolic processing \cite{moyer1967time,moyer1973comparing,dehaene1998abstract}, and is mostly attributed to a "mental number line" on which comparisons are done, such that it is easier to compare/encode distant points than close points \cite{dehaene1992varieties}.

\vspace*{1.5ex}\noindent\textbf{Comparison between animal sizes:}
We asked GPT-3 to compare the size of animals by completing a prompt of the following format: \\
\\
Q: Is [first stimulus] [smaller/bigger] than [second stimulus]? \\
A: \\

GPT-3 was queried for every two animals from an overall set, and for every possible combination (is first bigger than second, is second bigger than first, is first smaller than second, etc).

In order to get more data, we introduce variations in the query format. The question \& answer indicators (in the example above, Q\&A) could be any of the following: Q\&A, Question\&Answer, q\&a, question\&answer, QUESTION\&ANSWER. The separator between the indicator and the rest of text could be ``:'', ``)'', ``.'', `]'', ``\}'', ``;''. Each step was performed with all combinations (overall $5 \cdot 6 = 30$ combinations). In addition, each pair was tested with plural and singular forms. Overall, for each pair of animals, there were 2 combination of order, 2 of comparison word, 2 of plural/singular form, 30 of questioning way, giving 240 queries in total.

There were two sets of animals, the first with length of 3 letters each, and the second with length of 5 letters each: 
\begin{itemize}
\item \emph{First set}: ant, bat, owl, cat, pig, cow. 
\item \emph{Second set}: snail, raven, koala, camel, whale.
\end{itemize}
In order to prevent the confidence in pair comparisons from being influenced by the frequencies of the animals' names, which could obscure the distance effect, we selected animal names with similar frequencies.

For each set, two experiments were performed. The first was with the letters of the animals separated by a space (e.g., c o w) and the second with regular representation (e.g., cow). The idea was to test whether the effect could be presented with and without tokens.

We performed another experiment with animals taken from \cite{paivio1975perceptual} in which participants were asked to rate the sizes of various objects on a scale ranging from 1 to 10. We selected 7 animals of increasing size, with an average distance of 1-1.5 and a standard deviation smaller than 1. The animals are: 
\begin{itemize}
\item \emph{Pavio's set}: ant, rat, goose, wolf, donkey, bear, whale. 
\end{itemize}
However, since the names of the animals are not of fixed length, we didn't run the experiment with letters separated by spaces, as it might be that animals with longer names would be more confusing than animals with shorter names, thus obscuring the effect.

\vspace*{1.5ex}\noindent\textbf{Comparison between numbers:}
When asking GPT-3 to compare two numbers directly, it was too certain about the correct result, so no effect could be found. In order to overcome this, GPT-3 was asked to complete a text in the following format: \\
\\
x is [first number] \\
y is [second number] \\
Q: Is x [less/greater] than y? \\
A: \\

The same variations from the previous section were introduced here as well, except for the plural form, which is of course non-relevant for numbers. Overall, there were 120 variations for every two numbers. All numbers from one to nine were tested, in their wording form. In addition, another experiment was done with dozens (ten, twenty, etc) and hundreds (one hundred, two hundred, etc).

\vspace*{1.5ex}\noindent\textbf{Comparison between months:}
We asked GPT-3 to complete a text of the following format: \\
\\
Q: Is [first month] [before/after] [second month]? \\
A: \\

\noindent 
The same variations were used here, 120 for each two months. The first nine months were tested.

\vspace*{1.5ex}\noindent\textbf{Comparison between letters:}
We asked GPT-3 to complete a text of the following format: \\
\\
Q: In the alphabet, is [first letter] [before/after] [second letter]? \\
A: \\

The same variations were used here, 120 for each two letters. The first nine letters were tested.

\vspace*{1.5ex}\noindent\textbf{Results:}
As before, we define the confidence as the probability GPT-3 assigns to the right answer (yes/no), out of the overall probability GPT-3 assigns to "yes" and "no".

In the graphs, the $x$ axis is the distance between the compared stimuli, and the $y$ axis is the average confidence for such stimuli. It can be seen that, generally, as the distance increased, the confidence increased as well. We present only a few graphs, but the effect was similar for all animals/numbers comparisons. The full results are available on our github project.

\begin{figure}
\centerline{\includegraphics[height=2in]{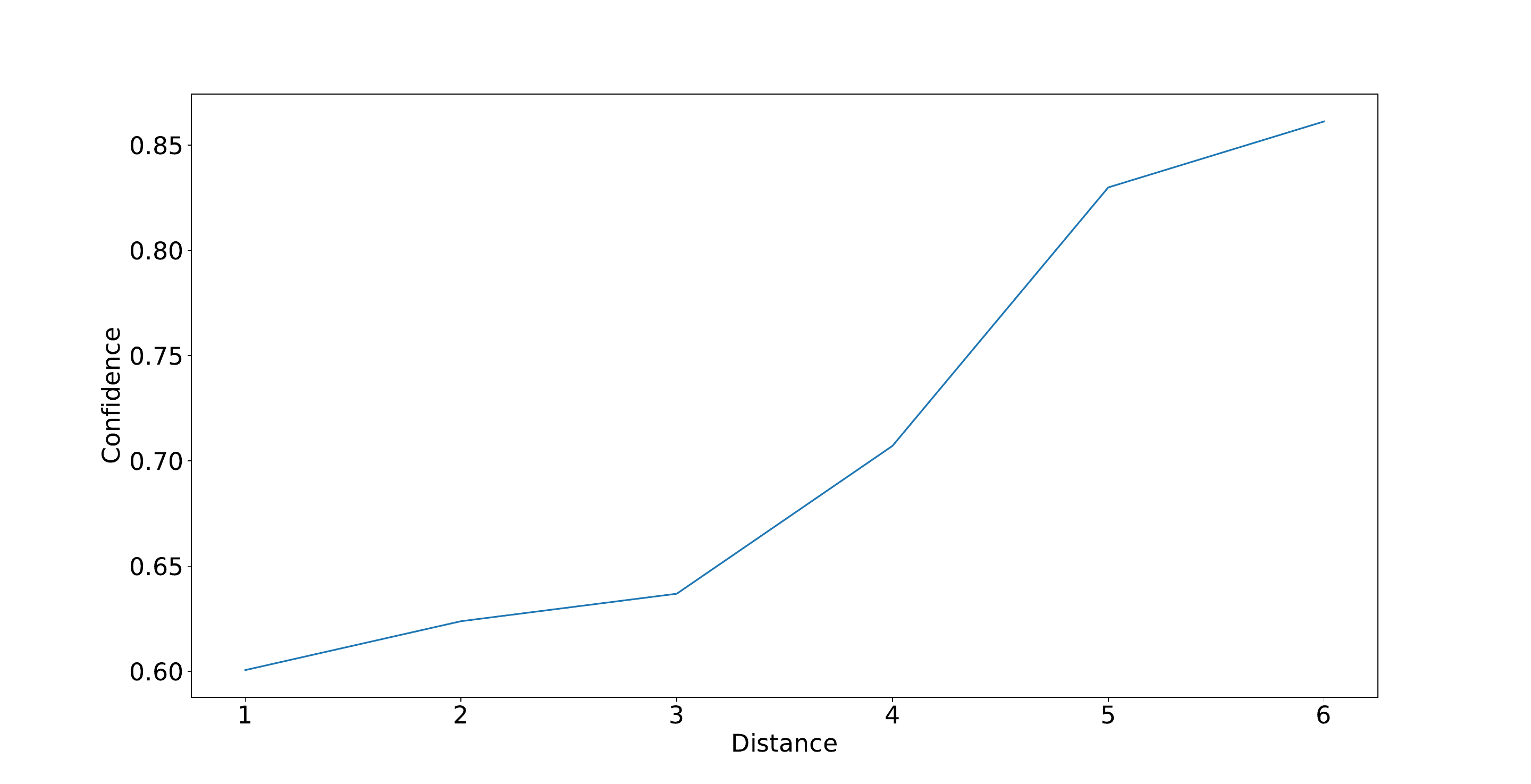}}
\caption{The distance effect with Pavio's animals. It can be seen that the confidence increased as the distance between the compared animals increased. ANOVA: $F(5, 5034)=39.45$, $p<0.001$, $MSE=0.16$.} \label{procstructfig}
\end{figure}

\begin{figure}
\centerline{\includegraphics[height=2in]{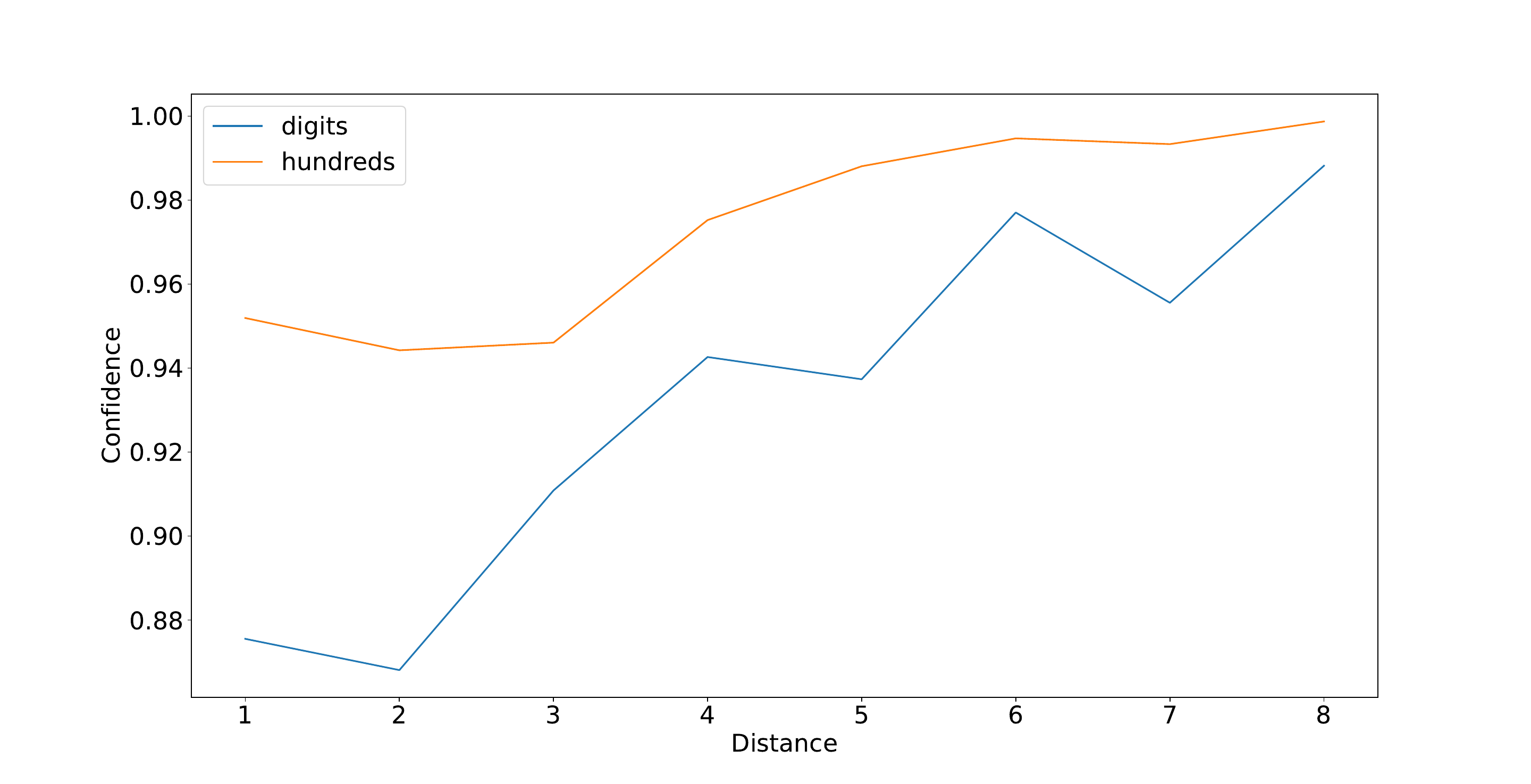}}
\caption{The distance effect with numbers. Not only the confidence increased as the distance increased, but merely adding the word 'hundred' to the stimuli increased GPT's confidence in the correct result. ANOVA for digits: $F(7, 4312)=15.06$, $p<0.01$, $MSE=0.06$. ANOVA for hundreds: $F(7, 4312)=10.59$, $p<0.01$, $MSE=0.02$.} \label{procstructfig}
\end{figure}

The effect, however, wasn't found for months and letters. Indeed, the effect was completely reversed for months: GPT-3 was more certain when comparing two following months than two distant months. Generally, the effect seems to be presented for size comparisons, but not for order comparisons.

\subsection{The SNARC Effect}

The SNARC effect refers to the association between left and small numbers, right and large numbers \cite{shaki2018deconstructing}. In a typical experiment, participants were required to respond to even numbers with their right hand and odd numbers with their left hand, or vice versa. Large numbers yielded faster responses with the right hand, and small numbers with the left \cite{dehaene1993mental}. The SNARC effect might originate from the directionality of the mental number line \cite{dehaene1992varieties} according to reading directions \cite{shaki2009reading} or alternatively could be attributed to associations between large numbers and good, good and right, and similarly between small, bad, and left \cite{proctor2006polarity} or even to the brain's asymmetry \cite{felisatti2020biological}.

Another version of SNARC, known as the vertical SNARC effect, is the association between small numbers with the downward direction and large numbers with the upward direction \cite{ito2004spatial}.

\vspace*{1.5ex}\noindent\textbf{Magnitude classification:}
We asked GPT-3 to complete a text of the following format (experiment 1): \\
\\
In the following instructions, X is equal to 5. \\
A word is about to be presented to you. \\
If the word represents a number [smaller/larger] than X, respond with "[left/right]". \\
If the word represents a number [larger/smaller] than X, respond with "[right/left]". \\
Word: [number word] \\
Response: \\

X could be any of the following: B, C, D, E, X, Y, W, Z.
\\
In another experiment (2), the following format was tested (i.e. implicit association test; see \cite{greenwald1998measuring, fischer2016measuring}): \\
\\
In the following instructions, X is equal to 5. \\
A word is about to be presented to you. \\
If the word is a spatial word associated with [right/left] or the word represents a number [smaller/larger] than X, respond with [first response]. \\
If the word is a spatial word associated with [left/right] or the word represents a number [larger/smaller] than X, respond with [second response]. \\
Word: [number word] \\
Response: \\

The responses were one of the following five characters: ``!'', ``@'', ``\#'', ``\$'', or ``\%'', and X could be either ``X'' or `Y''. All combinations were tested.

The possible number words were one, two, three, four, six, seven, eight and nine. All possible combinations (starting with the case of even number or the case of odd number, responding with right/left) were tested.

\vspace*{1.5ex}\noindent\textbf{Parity classification:}
We asked GPT-3 to complete a text of the following format (experiment 3): \\
\\
A word is about to be presented to you. \\
If the word represents an [even/odd] number, respond with "[right/left]". \\
If the word represents an [odd/even] number, respond with "[left/right]". \\
Word: [number word] \\
Response: \\

In another experiment (4), the following format was tested: \\
\\
A word is about to be presented to you. \\
If the word represents an [even/odd] number, respond with "[right/left]". \\
If the word represents an [odd/even] number, respond with "[left/right]". \\
Is the number greater than five? \\
Word: [number word] \\
Response: \\

And another variation (experiment 5): \\
\\
A word is about to be presented to you. \\
If the word represents an [even/odd] number, respond with "[right/left]". \\
If the word represents an [odd/even] number, respond with "[left/right]". \\
After responding, write whether or not the number is greater than five. \\
Word: [number word] \\
Response: \\

The separator between the word label and actual word, and between the response label and the actual response, could be any of the following: ``:'', ``;'', ``)'',``.'',``|'', ``]'', and ``\}'' and overall 7 possibilities.

\vspace*{1.5ex}\noindent\textbf{Vertical SNARC:}
All the above experiments were repeated, with down/up replacing left/right.

\vspace*{1.5ex}\noindent\textbf{Introducing uncertainty:}
Due to high certainty about the correct answer (approaching 1), we added spaces between the number words letters (e.g., o n e). We started by adding two spaces between each two adjacent letters and increased the spaces up to 20. When the confidence for a number word was less than $.6$, both when the desired response was congruent with the number and when it wasn't, we didn't test the number word with higher number of spaces.

\vspace*{1.5ex}\noindent\textbf{Results:}
The confidence is the probability GPT-3 assigns to correct response, out of the overall probability GPT-3 assigns to legal responses. In the results, for each number word, we present the average confidence when the query was congruent with the size of the numbers (small number with left, large with right) and the confidence when the query wasn't congruent (large number with left, small with right). We include only numbers of spaces for which at least one of the confidences (congruent/incongruent) was smaller than $.99$, and at least one was larger than $.6$.

Our results are summarised in the following table:

\begin{table}
\begin{center}
{\caption{The SNARC effect}\label{table1}}
\begin{tabular}{lccccccc}
\hline
\rule{0pt}{12pt}
%&\multicolumn{7}{c}{Processors}\\
%\cline{2-8}
\rule{0pt}{12pt}
Experiment&$\Diamond$&$\Box$&$\bigtriangleup$&$\bigcirc$&$\pentagon$&$\times$&
\\
\hline
\\[-6pt]
\quad 1 horizontal & .74 & .91 & <0.001 & -7.27 & 254 & 8 \\
\quad 1 vertical & .62 & .91 & <0.001 & -8.79 & 254 & 8 \\
\quad 2 horizontal & .82 & .91 & <0.001 & -4.52 & 382 & 8 \\
\quad 2 vertical & .85 & .87 & 0.3933 & -0.85 & 334 & 7 \\
\quad 3 horizontal & .82 & .68 & <0.001 & 3.47 & 222 & 8 \\
\quad 3 vertical & .71 & .65 & 0.1972 & 1.29 & 194 & 7 \\
\quad 4 horizontal & .45 & .89 & <0.001 & -10.89 & 194 & 7 \\
\quad 4 vertical & .38 & .84 & <0.001 & -10.89 & 222 & 8 \\
\quad 5 horizontal & .74 & .88 & <0.001 & -3.44 & 194 & 7 \\
\quad 5 vertical & .67 & .81 & 0.0021 & -3.13 & 194 & 7 \\
\\
\hline
\\[-6pt]
\multicolumn{8}{l}{
$\Diamond$ confidence with incongruence \ \
$\Box$ with congruence \ \
}
\\
\multicolumn{8}{l}{
$\bigtriangleup$ p-value \ \
$\bigcirc$ t-statistic \ \
$\pentagon$ degrees of freedom
}
\\
\multicolumn{8}{l}{
$\times$ analyzed digits
}
\end{tabular}
\end{center}
\end{table}

We hypothesise that the effect exists only when the size of the digits is relevant (all experiments, except 3). In some experiments, this was also the case with human participants \cite{shaki2018deconstructing}.

When the size was relevant, the effect seems to be present in all experiments except the case of 2-vertical (7/8 of the experiments).

\subsection{The Size Congruity Effect}

The size congruity effect refers to the phenomenon where people tend to respond faster when comparing the sizes of two stimuli that match in both their real and presentation sizes \cite{foltz1984mental,santens2011size}. For instance, when large animals' names are presented in a large font and small animals' names are presented in a small font, participants demonstrate a quicker response time than when the opposite is the case \cite{rubinsten2002ant}. A standard explanation is that the brain does both comparisons (presented and real sizes) and conflicting results reduce reaction time. Alternatively, it is harder to encode incongruent stimuli \cite{rubinsten2002ant}.

\vspace*{1.5ex}\noindent\textbf{Comparison between animals' sizes}
We asked GPT-3 to compare animals' sizes, by completing a prompt of the following format: \\
\\
Q: Is [first stimulus] [smaller/bigger] than [second stimulus]? \\
A: \\

Each time, one of the stimuli was capitalized (e.g., COW) and the other wasn't. Apart from that, the variations and the stimuli were the same as in the distance effect for animals.

In addition, the following set of animals was tested:
\begin{itemize}
\item \emph{4-animals}: Moth, frog, duck, goat, puma, bear.
\end{itemize}

\vspace*{1.5ex}\noindent\textbf{Comparison between numbers:}
GPT-3 was asked to compare numbers, by completing a prompt of the following format: \\
\\
{[first letter]} is [first number] \\
{[second letter]} is [second number] \\
Q: Is [first letter] [less/greater] than [second letter]? \\
A: \\

Each time one stimulus was capitalized (e.g., X is ONE, y is two). The same variations from the distance effect were used here.

Another experiment was done when the comparison words were [smaller/larger] instead (variation 2).

\vspace*{1.5ex}\noindent\textbf{Results:}
As before, the confidence is defined as the probability GPT-3 assigns to the right answer (yes/no), out of the overall probability GPT-3 assigns to ``yes'' and ``no''. When analyzing the results, we ignored pairs of animals that GPT-3 was certain about their sizes (more then $.99$ confidence, both for congruent and for incongruent queries) or too low confidence (less then $.6$, both for congruent and for incongruent queries).

Our results are as follows:

\begin{table}
\begin{center}
{\caption{The size congruity effect}\label{table1}}
\begin{tabular}{lccccccc}
\hline
\rule{0pt}{12pt}
Experiment&$\Diamond$&$\Box$&$\bigtriangleup$&$\bigcirc$&$\pentagon$&$\times$&
\\
\hline
\\[-6pt]
\quad Pavio's animals & .67 & .74 & <0.001 & -8.94 & 8158 & 17 \\
\quad 3-animals & .61 & .68 & <0.001 & -5.4 & 5758 & 12 \\
\quad 4-animals & .68 & .72 & 0.0239 & -2.2604 & 2398 & 5 \\
\quad 5-animals & .68 & .74 & <0.001 & -5.84 & 5758 & 12 \\
\quad spaced 3-animals & .63 & .67 & 0.0079 & -2.66 & 2878 & 6 \\
\quad spaced 4-animals & .64 & .70 & <0.001 & -5.65 & 4318 & 9 \\
\quad spaced 5-animals & .76 & .78 & 0.0096 & -2.59 & 5278 & 11 \\
\quad numbers (1) & .88 & .82 & <0.001 & 8.09 & 8638 & 36 \\
\quad numbers (2) & .79 & .83 & <0.001 & -5.18 & 8638 & 36 \\

\\
\hline
\\[-6pt]
\multicolumn{8}{l}{
$\Diamond$ confidence with incongruence \ \
$\Box$ with congruence \ \
}
\\
\multicolumn{8}{l}{
$\bigtriangleup$ p-value \ \
$\bigcirc$ t-statistic \ \
$\pentagon$ degrees of freedom
}
\\
\multicolumn{8}{l}{
$\times$ analyzed pairs
}
\end{tabular}
\end{center}
\end{table}

The impact was considerable for animals, but limited or even opposite when it comes to numbers.

\subsection{The Anchoring Effect}

The anchoring effect is a cognitive bias that occurs when people rely on an irrelevant piece of information they are exposed to (the ``anchor'') when making decisions or estimates. People tend to adjust their estimates or judgments from the initial anchor, but the adjustment is often insufficient, leading to biased decisions toward the original anchor \cite{tversky1974judgment,furnham2011literature}.

\vspace*{1.5ex}\noindent\textbf{Methodology:}
It is noteworthy that the anchoring effect manifests itself primarily when participants are unaware of the actual value being estimated; otherwise, they would simply provide the correct response, rendering any variance across anchors negligible (i.e., judgement under uncertainty: see \cite{kahneman1982judgment}). However, GPT-3 has very extensive knowledge and can readily provide accurate answers to diverse questions posed in conventional experiments (such as the number of countries in Africa), compelling us to seek an estimation task that we can be confident is outside its training set. To this end, we opted for measuring sequence lengths as the chosen task.

GPT-3 was asked to complete code of the following form (experiment 1): \\
\\
a = [anchor] \\
length = len('[sequence]') \# equals to \\

Or the following form (experiment 2): \\
\\
a = [first anchor] \\
b = [second anchor] \\
z = len('[sequence]') \# equals to \\

Here, the anchors are numbers, and sequence is a sequence of characters. We observed that GPT-3 can only estimate the length of sequences that are sufficiently long (over 30 characters), rather than providing an exact value. It should be noted that the characters are first converted into tokens before being processed by GPT, with some tokens representing multiple characters. To avoid that unexpected difficulty in the task, certain characters were chosen that do not form a token for sequences that do not contain any character presented twice in a row. The chosen characters are ``!'', ``\#'', ``\%'', ``\^'', ``\&'' and ``*'', and test sequences were randomly generated using these characters.

Two more variations were tested (experiments 3, 4): \\
\\
len('[anchor sequence]') \# equals to [anchor length] \\
len('[sequence]') \# equals to \\

\noindent And: \\
\\
len('[first anchor sequence]') \# equals to [first anchor length] \\
len('[second anchor sequence]') \# equals to [second anchor length] \\
len('[sequence]') \# equals to \\

Here the anchors exhibited greater similarity to the stimulus used in the actual task.

Throughout all stages of the experiments, anchors were classified into two categories: small and large. Small anchors were randomly selected from a uniform distribution within the range of 10 to 29, while large anchors were drawn from the range of 71 to 90. The evaluated sequence had a length between 40 to 60, and for each length, 20 queries were conducted with small anchors and 20 with large anchors, resulting in a total of 840 queries overall. For each query, the answer for which GPT-3 assigned the highest probability was taken.

\vspace*{1.5ex}\noindent\textbf{Results:}
Our results are summarised in the following table:
\begin{table}
\begin{center}
{\caption{The anchoring effect}\label{table1}}
\begin{tabular}{lccccccc}
\hline
\rule{0pt}{12pt}
Experiment&$\Diamond$&$\Box$&$\bigtriangleup$&$\bigcirc$&$\pentagon$&$\times$&
\\
\hline
\\[-6pt]
\quad 1 & 41.5 & 41.28 & 0.6102 &  0.5099 & 838 & 21 \\
\quad 2 & 41.08 & 39.90 & 0.0059 & 2.762 & 838 & 21 \\
\quad 3 & 45.78 & 42.04 & <0.001 & 9.58 & 838 & 21 \\
\quad 4 & 48.86 & 46.31 & <0.001 & 5.51 & 838 & 21 \\
\\
\hline
\\[-6pt]
\multicolumn{8}{l}{
$\Diamond$ estimation with small anchor \ \
$\Box$ with large anchor \ \
}
\\
\multicolumn{8}{l}{
$\bigtriangleup$ p-value \ \
$\bigcirc$ t-statistic \ \
$\pentagon$ degrees of freedom
}
\\
\multicolumn{8}{l}{
$\times$ analyzed lengths
}
\end{tabular}
\end{center}
\end{table}

We conclude that the effect doesn't appear to be presented. The mean estimates of the first two variations are similar when different anchors are used, while for the latter two, the effect is observed to be even reversed.

\section{Discussion}

We found that four out of five effects that we tested are presented by GPT-3: namely the priming, distance, SNARC, and size congruity effects. Although it is conceivable that some patterns in GPT-3 could perhaps exist due to pure randomness, we believe this result suggests more than a coincidence. 

To begin, we note that it is likely that GPT-3 was trained on papers describing the effects we have tested for. However, it is unlikely that these papers have text with formats similar to our queries, and it would be absurd to expect GPT-3 to generalize such as ``people respond faster to words recognition when words following associated words, and so I should assign such cases higher probabilities''. Hence, we conclude that GPT-3's knowledge about the effects is not the origin of their presentation within GPT-3 itself, and so we consider some alternative options.

\subsection{The Priming Effect}

The priming effect could be attributed, as with humans, to ``making the cognitive operations used to comprehend this content more accessible" when associated words follow the priming words \cite{janiszewski2014content}. Alternatively, we can say simply that associated words follow previous words more frequently than unrelated words and so GPT-3 handles it better. Apart from that, however, the effect is not likely to be imitated from people, as it's hard to see how this could be manifested in the training data.

\subsection{The Distance Effect}

As we saw, GPT-3's confidence increased as the distance between the stimuli it compared increased, for objects that included animals and numbers (similar results with RoBERTa found in \cite{talmor2020olmpics}). One may claim that for animals, the effect could be attributed to GPT-3 being merely unsure about their exact sizes but only having a vague, wide range estimation, which overlaps more for animals of closer sizes. However, it hardly explains the presence of the effect beyond the first few distances, and fails completely to explain the case for numbers, as without our added ``mental load" GPT-3 compares small numbers almost perfectly.
Another explanation was suggested by one of our group members, attributing the effect to the tokens' distance in the embedding space. Further experiments, however, showed it can't be the sole reason, as the effect remained when we presented the animals' letters separated by spaces, for which no tokens exist.

In contrast to priming, it is hard to explain the distance effect in terms of frequencies. Especially with numbers, it is somewhat surprising that GPT-3 is more confident when comparing distant numbers, as we expect that comparisons between consecutive numbers to be more frequent on the internet. Especially, we expect comparisons between digits to be more frequent than between hundreds.

To conclude, we attribute the effect to some kind of a "mental number line", as with humans  \cite{dehaene1992varieties}. It might be that it takes more time to decide about close stimuli, or to encode them in the first place \cite{van2008dissecting} \cite{van2011origins}. If GPT-3 indeed possess such a "mental number line", it's an indication for its ability to perform robust comparisons, even without direct training. Further investigations, such as measuring the frequencies of different numbers comparisons in the internet, might shed more light on the reasons for this effect being present with GPT-3.

\subsection{The SNARC Effect}

We showed that GPT-3 associates between left and small numbers, right and large numbers, and similarly with down and up. While the latter could be seen somewhat natural, attributed to graphs, floor numbers, and heights, the former is considered more arbitrary \cite{winter2015mental}. The lateralization of the brain, or reading directions might explain the effect with humans \cite{shaki2009reading,felisatti2020biological} but this explanation does not hold for GPT-3, as it lacks both (GPT-3 reads tokens one by one, without any spatial association regarding the process, similar to listening). The other explanation could be an association between left to bad, bad to small numbers, and between right-good-large, as some claims with humans \cite{proctor2006polarity}. Alternatively, GPT-3 may simply imitate this effect from humans, although it is hard to imagine how it is expressed in the training data. Future research may provide further insights.

\subsection{The Size Congruity Effect}

When comparing animals, GPT-3's confidence was higher when the smaller animal's names were presented with lowercase letters, and the larger with uppercase, compared to the reversed situation. Hence, it was demonstrating the size congruity effect. As with the other effects, it's hard to imagine this being expressed through training data.
 
Two major psychological models attempt to explain the effect. The shared decisions model claims that the decision about the presented sizes and the real sizes interfere in some way. In contrast, the shared representation model claims the encoding of the stimuli is more complicated when the stimuli are incongruent than when they congruent (e.g., encoded as small and large \emph{versus} presented-small-actually-large and presented-large-actually-small) \cite{rubinsten2002ant}. Either of these theories could account for our results, although it might be surprising that GPT-3 treats letters capitalization so similar to how people treat font sizes.

\section{Conclusions}

We have shown that some cognitive effects, namely the priming, distance, SNARC, and size congruity effect are presented in GPT-3, while the anchoring effect was absent. We have presented our methodology in detail, and some analogies that served us when turning real-world experiments to text-based. Finally we have speculated on the possible reasons why these effects are present in GPT-3, and provided some classical, psychological explanations to their existence in humans, which might be relevant to GPT-3 as well.

A huge body of work remains to be done on this and related problems. We have worked with GPT-3 -- it would be natural to study GPT-4 (released while this article was being written), and in particular to study whether the cognitive effects presented by GPT-3 are magnified or reduced in GPT-4. Comparison with LLMs outside OpenAI's GPT stable are also natural. Finally, note that the  question of \emph{methodology} for investigating questions like these with LLMs requires considerable further work.

\ack{We would like to thank Samuel Shaki for his useful insights from cognitive psychology.
The research was supported in part by the EU Project TAILOR under grant 952215.
}

\section*{Technical Appendix}

Our complete results, including the data and the code used to analyze it, are available on our github project: \href{https://github.com/GPTBiases/results}{https://github.com/GPTBiases/results}.

All experiments were conducted on GPT-3 and specifically the model \texttt{text-davinci-003}. In order to calculate the confidence we examined the 5 top probabilities predicted by GPT-3 (logprobs=5). Other settings, such as temperature or top-P, are not relevant since they determine how to sample a single token from the generated probability distribution, while we examined the distribution itself directly.
%\newpage
\bibliography{ecai}

\begin{thebibliography}{10}

\bibitem{almeida2019word}
Felipe Almeida and Geraldo Xex{\'e}o, `Word embeddings: A survey', {\em arXiv
  preprint arXiv:1901.09069}, (2019).

\bibitem{anderson1996architecture}
John~Robert Anderson, {\em The architecture of cognition}, volume~5, Psychology
  Press, 1996.

\bibitem{azariachatgpt}
Amos Azaria, `Chatgpt: More human-like than computer-like, but not necessarily
  in a good way', {\em The 45th Annual Meeting of the Cognitive Science
  Society}, (2023).

\bibitem{barsalou2014cognitive}
Lawrence~W Barsalou, `Cognitive psychology: An overview for cognitive
  scientists', (2014).

\bibitem{binz2023using}
Marcel Binz and Eric Schulz, `Using cognitive psychology to understand gpt-3',
  {\em Proceedings of the National Academy of Sciences}, {\bf 120}(6),
  e2218523120, (2023).

\bibitem{dehaene1992varieties}
Stanislas Dehaene, `Varieties of numerical abilities', {\em Cognition}, {\bf
  44}(1-2),  1--42, (1992).

\bibitem{dehaene1993mental}
Stanislas Dehaene, Serge Bossini, and Pascal Giraux, `The mental representation
  of parity and number magnitude.', {\em Journal of experimental psychology:
  General}, {\bf 122}(3),  371, (1993).

\bibitem{dehaene1998abstract}
Stanislas Dehaene, Ghislaine Dehaene-Lambertz, and Laurent Cohen, `Abstract
  representations of numbers in the animal and human brain', {\em Trends in
  neurosciences}, {\bf 21}(8),  355--361, (1998).

\bibitem{felisatti2020biological}
Arianna Felisatti, Jochen Laubrock, Samuel Shaki, and Martin~H Fischer, `A
  biological foundation for spatial--numerical associations: the brain's
  asymmetric frequency tuning', {\em Annals of the New York Academy of
  Sciences}, {\bf 1477}(1),  44--53, (2020).

\bibitem{fischer2016measuring}
Martin~H Fischer and Samuel Shaki, `Measuring spatial--numerical associations:
  Evidence for a purely conceptual link', {\em Psychological Research}, {\bf
  80},  109--112, (2016).

\bibitem{floridi2020gpt}
Luciano Floridi and Massimo Chiriatti, `Gpt-3: Its nature, scope, limits, and
  consequences', {\em Minds and Machines}, {\bf 30},  681--694, (2020).

\bibitem{foltz1984mental}
Gregory~S Foltz, Steven~E Poltrock, and George~R Potts, `Mental comparison of
  size and magnitude: size congruity effects.', {\em Journal of Experimental
  Psychology: Learning, Memory, and Cognition}, {\bf 10}(3),  442, (1984).

\bibitem{furnham2011literature}
Adrian Furnham and Hua~Chu Boo, `A literature review of the anchoring effect',
  {\em The journal of socio-economics}, {\bf 40}(1),  35--42, (2011).

\bibitem{greenwald1998measuring}
Anthony~G Greenwald, Debbie~E McGhee, and Jordan~LK Schwartz, `Measuring
  individual differences in implicit cognition: the implicit association
  test.', {\em Journal of personality and social psychology}, {\bf 74}(6),
  1464, (1998).

\bibitem{hutchison2013semantic}
Keith~A Hutchison, David~A Balota, James~H Neely, Michael~J Cortese, Emily~R
  Cohen-Shikora, Chi-Shing Tse, Melvin~J Yap, Jesse~J Bengson, Dale Niemeyer,
  and Erin Buchanan, `The semantic priming project', {\em Behavior research
  methods}, {\bf 45},  1099--1114, (2013).

\bibitem{ito2004spatial}
Yasuhiro Ito and Takeshi Hatta, `Spatial structure of quantitative
  representation of numbers: Evidence from the snarc effect', {\em Memory \&
  Cognition}, {\bf 32},  662--673, (2004).

\bibitem{janiszewski2014content}
Chris Janiszewski and Robert~S Wyer~Jr, `Content and process priming: A
  review', {\em Journal of consumer psychology}, {\bf 24}(1),  96--118, (2014).

\bibitem{jones2022capturing}
Erik Jones and Jacob Steinhardt, `Capturing failures of large language models
  via human cognitive biases', {\em arXiv preprint arXiv:2202.12299}, (2022).

\bibitem{kahneman1982judgment}
Daniel Kahneman, Stewart~Paul Slovic, Paul Slovic, and Amos Tversky, {\em
  Judgment under uncertainty: Heuristics and biases}, Cambridge university
  press, 1982.

\bibitem{lin2022survey}
Tianyang Lin, Yuxin Wang, Xiangyang Liu, and Xipeng Qiu, `A survey of
  transformers', {\em AI Open}, (2022).

\bibitem{moyer1973comparing}
Robert~S Moyer, `Comparing objects in memory: Evidence suggesting an internal
  psychophysics', {\em Perception \& Psychophysics}, {\bf 13},  180--184,
  (1973).

\bibitem{moyer1967time}
Robert~S Moyer and Thomas~K Landauer, `Time required for judgements of
  numerical inequality', {\em Nature}, {\bf 215}(5109),  1519--1520, (1967).

\bibitem{paivio1975perceptual}
Allan Paivio, `Perceptual comparisons through the mind’s eye', {\em Memory \&
  Cognition}, {\bf 3}(6),  635--647, (1975).

\bibitem{park2023artificial}
Peter~S Park, Philipp Schoenegger, and Chongyang Zhu, `Artificial intelligence
  in psychology research', {\em arXiv preprint arXiv:2302.07267}, (2023).

\bibitem{posner2005timing}
Michael~I Posner, `Timing the brain: Mental chronometry as a tool in
  neuroscience', {\em PLoS biology}, {\bf 3}(2),  e51, (2005).

\bibitem{proctor2006polarity}
Robert~W Proctor and Yang~Seok Cho, `Polarity correspondence: A general
  principle for performance of speeded binary classification tasks.', {\em
  Psychological bulletin}, {\bf 132}(3),  416, (2006).

\bibitem{radford2018improving}
Alec Radford, Karthik Narasimhan, Tim Salimans, Ilya Sutskever, et~al.,
  `Improving language understanding by generative pre-training', (2018).

\bibitem{roediger1987effects}
Henry~L Roediger and Teresa~A Blaxton, `Effects of varying modality, surface
  features, and retention interval on priming in word-fragment completion',
  {\em Memory \& cognition}, {\bf 15}(5),  379--388, (1987).

\bibitem{rubinsten2002ant}
Orly Rubinsten and Avishai Henik, `Is an ant larger than a lion?', {\em Acta
  psychologica}, {\bf 111}(1),  141--154, (2002).

\bibitem{santens2011size}
Seppe Santens and Tom Verguts, `The size congruity effect: Is bigger always
  more?', {\em Cognition}, {\bf 118}(1),  94--110, (2011).

\bibitem{shaki2018deconstructing}
Samuel Shaki and Martin~H Fischer, `Deconstructing spatial-numerical
  associations', {\em Cognition}, {\bf 175},  109--113, (2018).

\bibitem{shaki2009reading}
Samuel Shaki, Martin~H Fischer, and William~M Petrusic, `Reading habits for
  both words and numbers contribute to the snarc effect', {\em Psychonomic
  bulletin \& review}, {\bf 16}(2),  328--331, (2009).

\bibitem{sperber1979semantic}
Richard~D Sperber, Charley McCauley, Ronnie~D Ragain, and Carolyne~M Weil,
  `Semantic priming effects on picture and word processing', {\em Memory \&
  Cognition}, {\bf 7}(5),  339--345, (1979).

\bibitem{talmor2020olmpics}
Alon Talmor, Yanai Elazar, Yoav Goldberg, and Jonathan Berant, `olmpics-on what
  language model pre-training captures', {\em Transactions of the Association
  for Computational Linguistics}, {\bf 8},  743--758, (2020).

\bibitem{tversky1974judgment}
Amos Tversky and Daniel Kahneman, `Judgment under uncertainty: Heuristics and
  biases: Biases in judgments reveal some heuristics of thinking under
  uncertainty.', {\em science}, {\bf 185}(4157),  1124--1131, (1974).

\bibitem{van2008dissecting}
Filip Van~Opstal, Wim Gevers, Wendy De~Moor, and Tom Verguts, `Dissecting the
  symbolic distance effect: Comparison and priming effects in numerical and
  nonnumerical orders', {\em Psychonomic Bulletin \& Review}, {\bf 15},
  419--425, (2008).

\bibitem{van2011origins}
Filip Van~Opstal and Tom Verguts, `The origins of the numerical distance
  effect: The same--different task', {\em Journal of Cognitive Psychology},
  {\bf 23}(1),  112--120, (2011).

\bibitem{winter2015mental}
Bodo Winter, Teenie Matlock, Samuel Shaki, and Martin~H Fischer, `Mental number
  space in three dimensions', {\em Neuroscience \& Biobehavioral Reviews}, {\bf
  57},  209--219, (2015).

\end{thebibliography}

\end{document}